\documentclass{ngsm2024} %

\usepackage{booktabs}
\usepackage{wrapfig}
\usepackage{comment}

\title[State Soup]{State Soup: In-Context Skill Learning, Retrieval and Mixing}

\optauthor{%
\Name{Maciej Pióro$^*$} \Email{maciej.pioro@ideas-ncbr.pl}\\
\addr IDEAS NCBR, IPPT PAN
\AND
\Name{Maciej Wołczyk$^*$} \Email{maciej.wolczyk@ideas-ncbr.pl}\\
\addr IDEAS NCBR
\AND
\Name{Razvan Pascanu} \Email{razp@google.com}\\
\addr Google Deepmind
\AND
\Name{Johannes von Oswald$^\odot$} \Email{jvoswald@google.com}\\
\Name{Jo\~ao Sacramento$^\odot$} \Email{joaosacramento@google.com}\\
\addr Google, Paradigms of Intelligence Team
}

\begin{document}

\maketitle

\def\footnoteseptext{ }
\def\thefootnote{$^*$}\footnotetext{Equal Contribution}\def\thefootnote{\arabic{footnote}}
\def\thefootnote{$^\odot$}\footnotetext{Co-Senior Authorship}\def\thefootnote{\arabic{footnote}}
\def\footnoteseptext{. }

\begin{abstract}
  A new breed of gated-linear recurrent neural networks has reached state-of-the-art performance on a range of sequence modeling problems. Such models naturally handle long sequences efficiently, as the cost of processing a new input is independent of sequence length. 
  Here, we explore another advantage of these stateful sequence models, inspired by the success of model merging through parameter interpolation. Building on parallels between fine-tuning and in-context learning, we investigate whether we can treat internal states as task vectors that can be stored, retrieved, and then linearly combined, exploiting the linearity of recurrence. We study this form of fast model merging on Mamba-2.8b, a pretrained recurrent model, and present preliminary evidence that simple linear state interpolation methods suffice to improve next-token perplexity as well as downstream in-context learning task performance.
\end{abstract}

\section{Introduction}
Transformers \cite{vaswani_attention_2017} have become the standard neural network architecture for sequence modeling. Their parallelizable training and predictable scaling behavior \cite{kaplan_scaling_2020} have led to unprecedented performance in a wide range of problem domains, with language modeling being the prime example. However, this architecture comes with the drawback that the memory and computational costs of inference scale quadratically with context length. This undesirable property has led to continued interest in recurrent models that propagate forward an internal state as a sequence is processed, for which inference costs instead scale linearly with context length.

Recently, great strides have been made in recurrent neural network (RNN) architecture research \cite[see, e.g.,][]{gu_efficiently_2022,orvieto_resurrecting_2023,gu_mamba_2023,yang_gated_2023,de_griffin_2024,beck_xlstm_2024}. We now have a number of architectures that can be trained as efficiently as Transformers, some of which exhibit similar scaling laws in language modeling up to the billion-parameter range \cite{gu_mamba_2023,yang_gated_2023,de_griffin_2024,beck_xlstm_2024}. At the core of these modern RNNs -- and of particular importance to the present paper -- is the use of gated-linear recurrences, which marry RNN gating techniques \cite{hochreiter_long_1997,cho_properties_2014} with classical linear state-space models, developed within control and linear filter theory. Linear recurrences enable computationally-efficient training (in particular for earlier ungated variants) and lead to good optimization properties when correctly parametrized \cite{gu_efficiently_2022,smith_simplified_2023,orvieto_resurrecting_2023}.

Here, we further exploit the linearity of modern RNNs and introduce \emph{state soups}. Inspired by model soups \citep{von_oswald_neural_2021,wortsman_model_2022,rame_rewarded_2024}, which improve or change the training objective by linearly interpolating the parameters of several models, we propose instead to perform state-space interpolations. As we show below, linear interpolation is exact for a single gated-linear recurrent layer, and approximately correct in a number of experiments performed with Mamba 2.8B, a pretrained modern RNN comprising dozens of layers. Conceptually, state soups not only enable parallel information processing across independent models, but also the caching of preprocessed information that can be later retrieved to augment a novel query, reminiscent of retrieval-augmented generation  \cite{lewis_retrieval-augmented_2020} and complementary learning systems theory \cite{mcclelland_why_1995,kumaran_what_2016}. We demonstrate the latter on a recently developed suite of in-context learning tasks \cite{todd_function_2024}, finding that query-based retrieval over a library of in-context learned skills is possible without any model fine-tuning. Finally, we find a few successful instances where a form of linear task arithmetic is possible, similarly to recent work on function vectors in Transformers \cite{todd_function_2024,hendel_-context_2023} and to the seminal discovery of word vector arithmetics \cite{mikolov_distributed_2013}.

\section{State soup} 

We explore the idea of building a library of in-context learned (ICL) skills, represented by RNN states that can be used for retrieval and mixing. For this purpose, we employ a set of simple ICL tasks proposed by~\cite{todd_function_2024}, where a single task is a sequence of $k$ examples formatted as (\texttt{<question> $\rightarrow $<answer>\textbackslash n}). We use a pretrained Mamba~\cite{gu_mamba_2023} model with 2.8B parameters for generating the skills for the library and subsequent testing.
Each skill-representing state is obtained by processing $32$ examples. For each task, we sample multiple disjoint sets of examples, so that we have multiple RNN states for each skill. For more technical details, see Appendix~\ref{app:technical}.

Using this setup, we ask three questions. (1)  \textbf{Task retrieval}: Given a short task example, can we retrieve the corresponding state from the skillset? (2) \textbf{State mixing}: Can we mix different states to boost results? 
(3) \textbf{State mixing with sequential data}: Can we apply mixing to sequential data?

\begin{figure}
\includegraphics[width=0.49\textwidth]{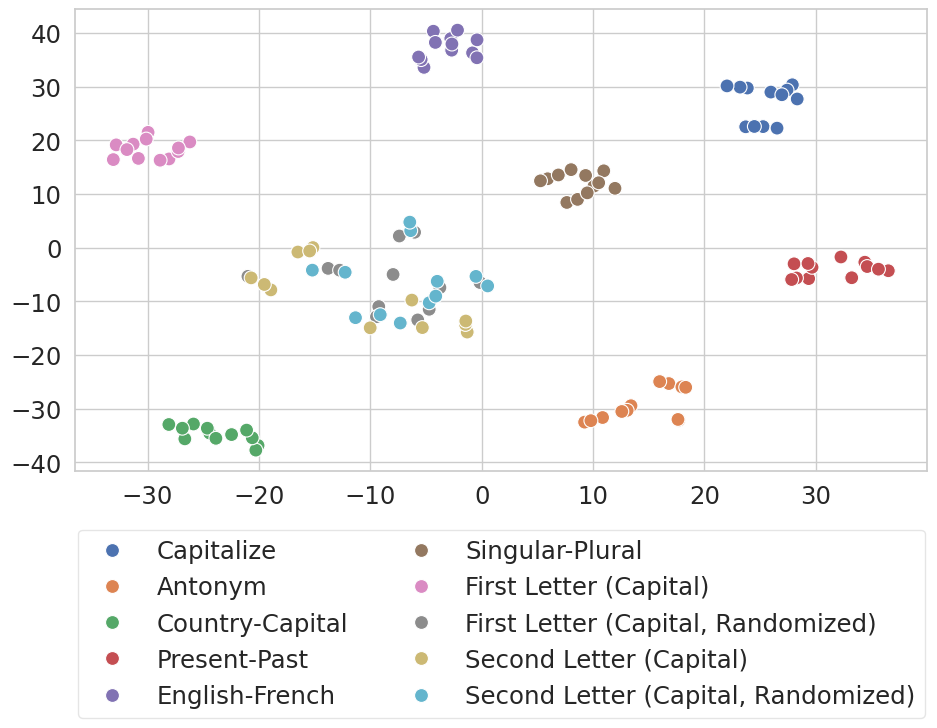}
\includegraphics[width=0.49\textwidth]{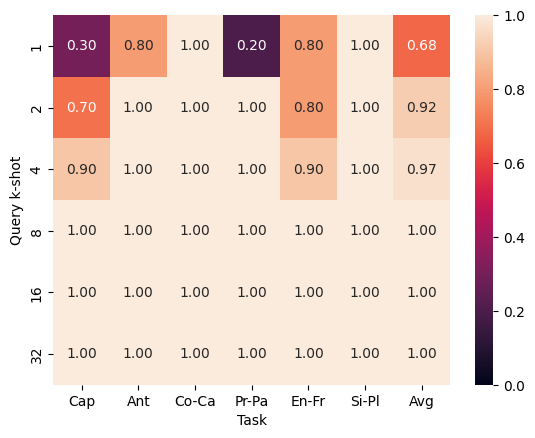}

\caption{(Left) In our skill library, states from the same task are clustered under T-SNE projection. (Right) Using a query state obtained after processing $k$ examples from a given task (y-axis), we check the probability that the closest state in the library is from the same task.}
    \label{fig:kshot_retrieval}
\vspace{-0.7cm}
\end{figure}

\subsection{Task retrieval}
\label{sec:retrieval}

In Figure~\ref{fig:kshot_retrieval} (left), we take an intermediate layer\footnote{We observed that intermediate layers encode the task most reliably, so we use the $32$nd layer our of $64$ in retrieval experiments.} from every state in our skill library, and we project it to two dimensions using T-SNE. As a result, we obtain a proper clustering, where states corresponding to the same task are grouped together. Even more interestingly, tasks that Mamba cannot learn are also clustered together, suggesting that the task essence can be easily decoded from the state vector.

To quantitatively test the retrieval capabilities of the model, we verify if a query vector $q^k_\tau$ obtained by processing a $k$-shot example from a task $\tau$ from the skill library can be correctly identified as coming from $\tau$. In particular, in Figure~\ref{fig:kshot_retrieval} (right), we check if the closest neighbor of $q^k_\tau$ is also a $\tau$-state. We check different values of $k$, and we we observe that our approach is able to reliably identify the correct task even for $k$ as low as $2$-shots, while for some tasks, it is possible even with $1$-shot.

\subsection{State mixing}
\label{sec:mixing}
\begin{figure}
\centering
    \includegraphics[width=\textwidth]{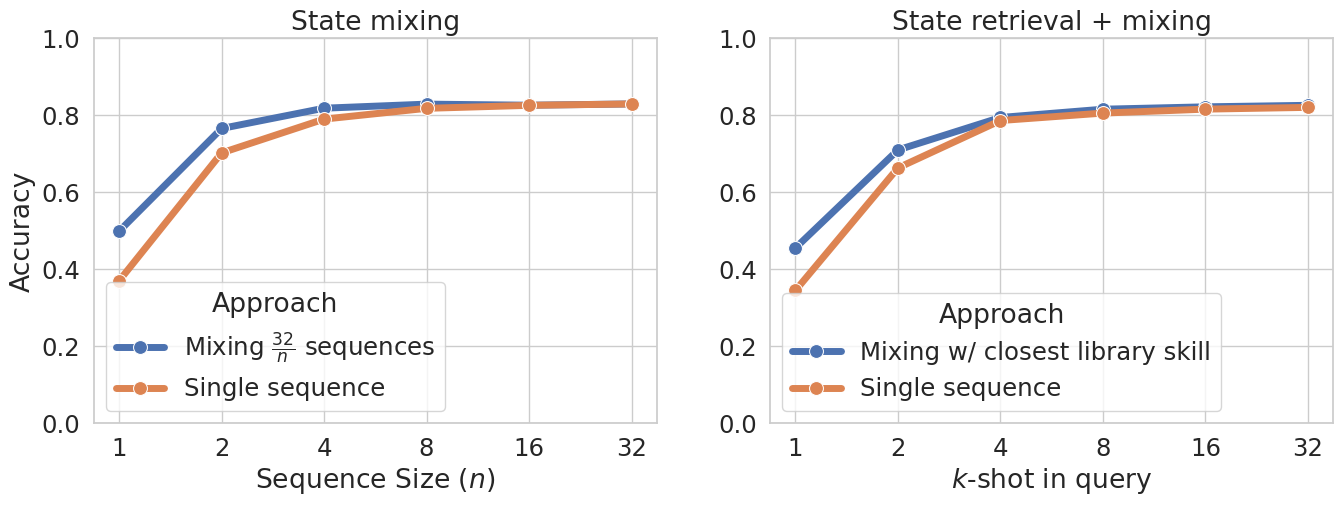}
    \caption{State retrieval and mixing improves few-shot learning performance. (Left) The x-axis represents the number of examples in the processed sequence. (Right) The x-axis represents the number of examples observed in the query state.} 
    \label{fig:icl}
    \vspace{-0.9cm}
\end{figure}

Having established that relevant-state retrieval can be easily accomplished, we turn to state mixing, with the goal of using a task library to enhance the few-shot performance of our model. To mix the states, we simply take a mean over them.
In our first experiment, see Figure \ref{fig:icl} (left), we compare the standard in-context learning with mixing. In the baseline setting (orange line) Mamba sees a single sequence of $k$ examples, where $k$ is depicted on the $x$ axis. On the other hand, our simple mixing strategy (blue line) mixes $\frac{32}{k}$ states, each obtained by independently processing $k$ examples. As such, mixing always uses $32$ examples in total. We observe that although mixing $32$ $1$-shot states does not lead to great results, mixtures of $4$-shot states are already on par or even slightly better than processing the whole $32$-shot at once.

With positive results from the first experiment, we can move to incorporating states retrieved from the state library. In this setup, each query state $q_\tau^k$ is obtained by processing $k$-shot demonstrations from task $\tau$. We use $q_\tau^k$ to retrieve the most similar state from the library, which is then mixed with the query state. The mixed state is finally used as the initial state for processing the test sample. The results shown in Figure~\ref{fig:icl} (right) show that we can boost the performance, especially with small $k$

Additionally, we do preliminary tests of mixing states from distinct tasks. In particular, we find that mixing states resulting from the "counting" task (\texttt{one $\rightarrow$ two $\rightarrow$ \dots $\rightarrow$ fourteen $\rightarrow$}) and "English-French" translation task resulted in the model predicting the next number in French (\texttt{quinze}). Note that obtaining this result required taking a weighted mean of the two states and is not yet backed by quantitative studies. However, we deem this an intriguing research direction.

\subsection{Mixing with sequential data}
\label{sec:sequential_mixing}

In our previous experiments, the ordering of the mixed chunks was irrelevant. However, the data is inherently ordered in many practical scenarios, such as processing long sequences. Here, we propose a mixing strategy that takes the sequential nature of data into consideration.

A single discretized SSM layer that processes the sequence $x_1, \ldots, x_t$ can be depicted recursively as: $f(x_1, \ldots, x_t) = A_t f(x_1, \ldots, x_{t-1}) + B_t x_t$, see Appendix~\ref{app:technical} for a detailed notation. Observe that due to the linearity of SSM, for each $k \in \{1, \ldots t-1\}$  we can write this equation as:
$f(x_1, \ldots, x_t) = f(x_{k+1}, \ldots, x_t) + \left ( \prod_{k'=1}^{t-k}  A_{k'} \right ) f(x_1, \ldots, x_k)$.

As such, in the linear setting, we can independently process sequences $x_1, \ldots, x_k$ and $x_{k+1}, \ldots, x_t$ and combine them exactly as long as we have the $\prod_{k} A_{k'}$ matrix. Fortunately, this matrix is computed for the parallel scan algorithm that enables efficient training of different SSM architectures~\cite{smith_simplified_2023}. We call this approach \textit{A-decay mixing}. %

\begin{wrapfigure}[12]{r}{0.45\textwidth}
\vspace{-0.9cm}
    \centering
    \includegraphics[width=\linewidth]{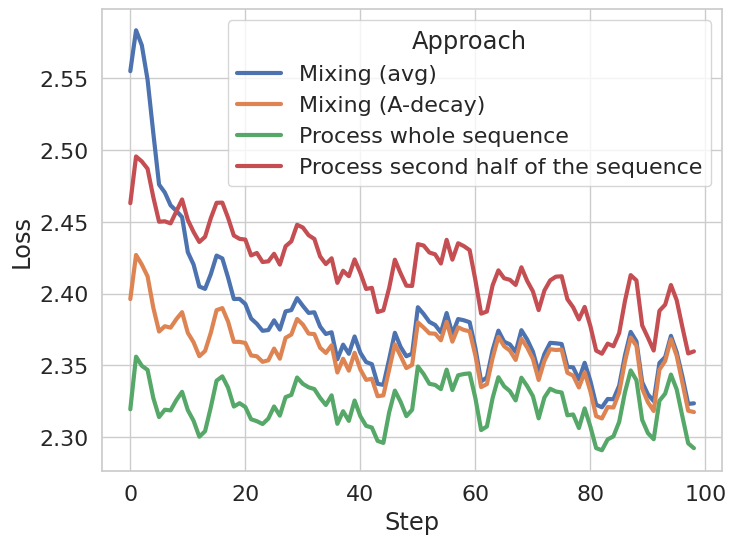}
   \vspace{-1.1cm} 
    \caption{Different mixing approaches on the next token prediction problem.}
    \label{fig:next_token_pred}
\end{wrapfigure}

For a single SSM layer, A-decay mixing gives us exactly the same solution as processing the whole sequence. However, in the full Mamba architecture, the inputs to the subsequent SSM layers depend on the outputs of the previous layer, and as such, A-decay mixing will only give us an approximation. 
To empirically test its quality, we take $10000$ sequences from the realnewslike dataset~\cite{raffel2020exploring}, and we divide each of them into three chunks: $c_1, c_2, c_{\text{test}}$, each with $100$ tokens. We check the prediction loss on $c_{\text{test}}$ using different RNN states: (1) sequential processing of $c_1$ and $c_2$, (2) sequential processing of $c_2$, (3) mean-mixing of states after processing separately $c_1$ and $c_2$, (4) same as previous but with A-decay mixing. Figure~\ref{fig:next_token_pred} shows that the A-decay mixing outperforms both mean-mixing and starting from a state that only processed $c_2$. However, the performance is still worse than the model that saw both $c_1$ and $c_2$, suggesting that A-decay offers us only an approximate solution in cases when the RNN layers are stacked.

\section{Conclusion}

Neural systems that combine some form of volatile fast learning with persistent memory stores have been previously studied in meta-learning \cite[e.g.,][]{santoro_meta-learning_2016}. Our preliminary findings suggest that it is possible to leverage the in-context learning abilities of recurrent neural networks to generate task representations that can be committed to memory and then later retrieved and reused or even repurposed. As the capacity to learn and compress long \cite[`many-shot';][]{agarwal_many-shot_2024} tasks in-context with RNNs increases, so will the relative advantage of our method against Transformer-based alternatives, which do not offer a way to reuse preprocessed states off-the-shelf at $\mathcal{O}(1)$ cost (but see \citep{todd_function_2024,hendel_-context_2023}).  In future work, we wish to extend our setup to include broader and more realistic tasks, and to quantitatively analyze the ability to perform task arithmetics.

\section*{Acknowledgments}
MP's and MW's work was funded by IDEAS NCBR. The research was supported by PL-Grid infrastructure (grant PLG/2024/017060). We also benefited from the Entropy cluster (hosted at the Faculty of Mathematics, Informatics and Mechanics of the University of Warsaw) funded by NVIDIA, Intel, the Polish National Science Center grant 2022/45/N/ST6/02222, and ERC Starting Grant TOTAL. We would like to thank \href{https://writer.com/}{Writer} who also supported us with computational resources. The authors want to thank Seijin Kobayashi and Xu He for valuable discussions throughout the project and notes on the manuscript.

\bibliography{state-soup}
\appendix

\section{Technical details}
\label{app:technical}
In all of our experiments we use the Huggingface Mamba-2.8b implementation which utilizes the GPT-NeoX-20B tokenizer~\cite{black2022gpt}.
The basis of our experiments in sections \ref{sec:retrieval} and \ref{sec:mixing} are the six main tasks introduced and investigated in \cite{todd_function_2024}, namely \textbf{Ant}onym (flawed: perfect, unrelated: related), \textbf{Cap}italize (lift: Lift, mirror: Mirror), \textbf{Co}untry\textbf{-Ca}pital (Egypt: Cairo, Poland: Warsaw), \mbox{\textbf{En}glish\textbf{-Fr}ench} (satisfy: satisfaire, here: ici), \mbox{\textbf{Pr}esent\textbf{-Pa}st} (assist: assisted, speak: spoke) and \textbf{Si}ngular\textbf{-Pl}ural (bat: bats, mouse: mice). 

In the clustering experiments, we use two further tasks, First Letter, Capital (python: P, finch: F) and Second Letter, Capital (ecstatic: C, change: H) along with \textit{Randomized} variations created by randomly shuffling the labels. This preserves the label distribution but removes any relationship between the input and the output.

\subsection{Task Retrieval}
In the clustering experiments, we create 12 16-shot demonstrations per task, while ensuring that no sample appears in multiple demonstrations. We then perform t-SNE dimensionality reduction \cite{vandermaaten_tsne} using the t-SNE-CUDA \cite{chan2019gpu_tsne} library.

In the retrieval experiments, we create a library of 10 states per task, with each state resulting from processing 32-shot demonstration.
For $k=1, 2, 4, \dots, 32$ we create 10 $k$-shot demonstrations, ensuring no overlap with the library. To measure the similarity of the queries to the states in the library, we use the cosine similarity of the SSM state at the 32nd layer.

\subsection{State Mixing}
In the state mixing experiments (Figure \ref{fig:icl}, left), we select 500 random test samples from the datasets, apart from the Country-Capital, Present-Past and Singular-Plural tasks, which only contain 197, 293 and 205 samples respectively - in those cases, we perform testing on all samples. We ensure that the test sample is not present in the few-shot samples or the samples used to obtain the mixed state.

In the "retrieval + mixing" experiments, we first randomly divide each dataset into two halves and only use samples from one of the halves to create the library, while testing the performance on the other half. Once again, we use 500 samples if possible, however, due to the halving, the Capitalize dataset is another dataset for which we use a smaller number (407) of test samples.

\subsection{Mixing with sequential data}

Here, we provide an explanation for the notation used in Section~\ref{sec:sequential_mixing}. 
The discretized SSM equation can be written as:
\begin{equation}
    f(x_1, \ldots, x_t) = A_t f(x_1, \ldots, x_{t-1}) + B_t x_t,
\end{equation}
where $f$ represents the SSM function, $x_1, ..., x_t$ are the tokens to be processed and each token is a vector of dimensionality $D$. $A_t \in \mathbb{R}^{H \times H}$ and $B_t \in \mathbb{R}^{H \times D}$, are the parameters of the model, where $H$ is the internal (hidden) dimensionality of the RNN. Both $A_t$ and $B_t$ might be time- and input-dependent, as in Mamba~\cite{gu_mamba_2023}, or independent as in the original SSMs~\cite{gu_efficiently_2022}.

\section{Additional results}
\subsection{State Retrieval}
\begin{figure}
\centering
    \includegraphics[width=\textwidth]{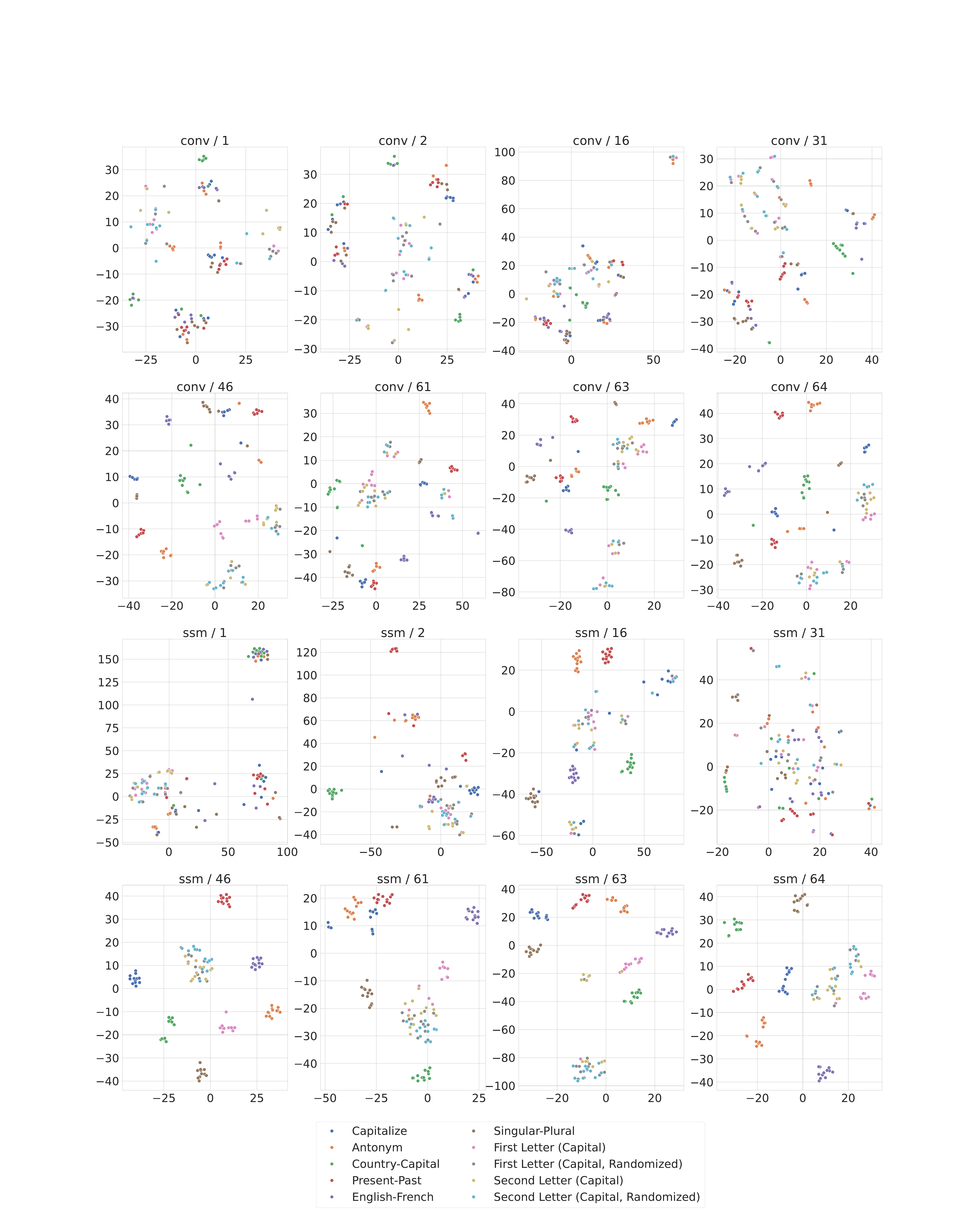}
    \caption{States clustering - dimensionality reduction performed with t-SNE. Both SSM and convolution states are investigated.} 
    \label{fig:clustering_grid_tsne}
\end{figure}

\begin{figure}
\centering
    \includegraphics[width=\textwidth]{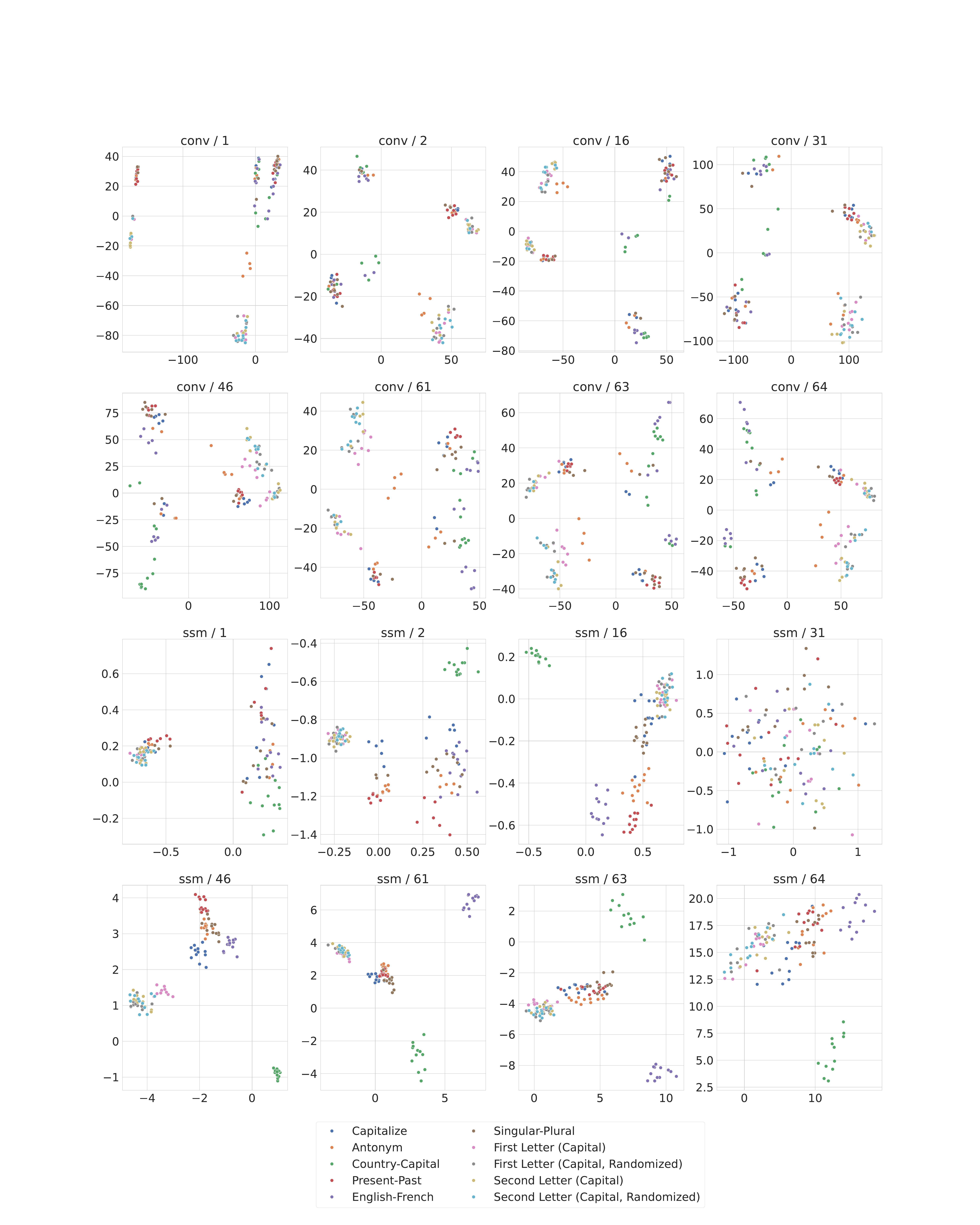}
    \caption{States clustering - dimensionality reduction performed with PCA. Both SSM and convolution states are investigated.} 
    \label{fig:clustering_grid_pca}
\end{figure}

The clustering of states belonging to the same task (or lack thereof) after dimensionality reduction can be observed across the layers of the model. We can also vary whether we explore the model's SSM or the convolutional state. The dimensionality reduction method plays a crucial role, with t-SNE producing much more visible and interpretable patterns than PCA. The plots showing the 2-dimensional visualizations of the states are presented in Figures \ref{fig:clustering_grid_tsne} and \ref{fig:clustering_grid_pca}.

\subsection{State Mixing}
\begin{figure}
\centering
    \includegraphics[width=\textwidth]{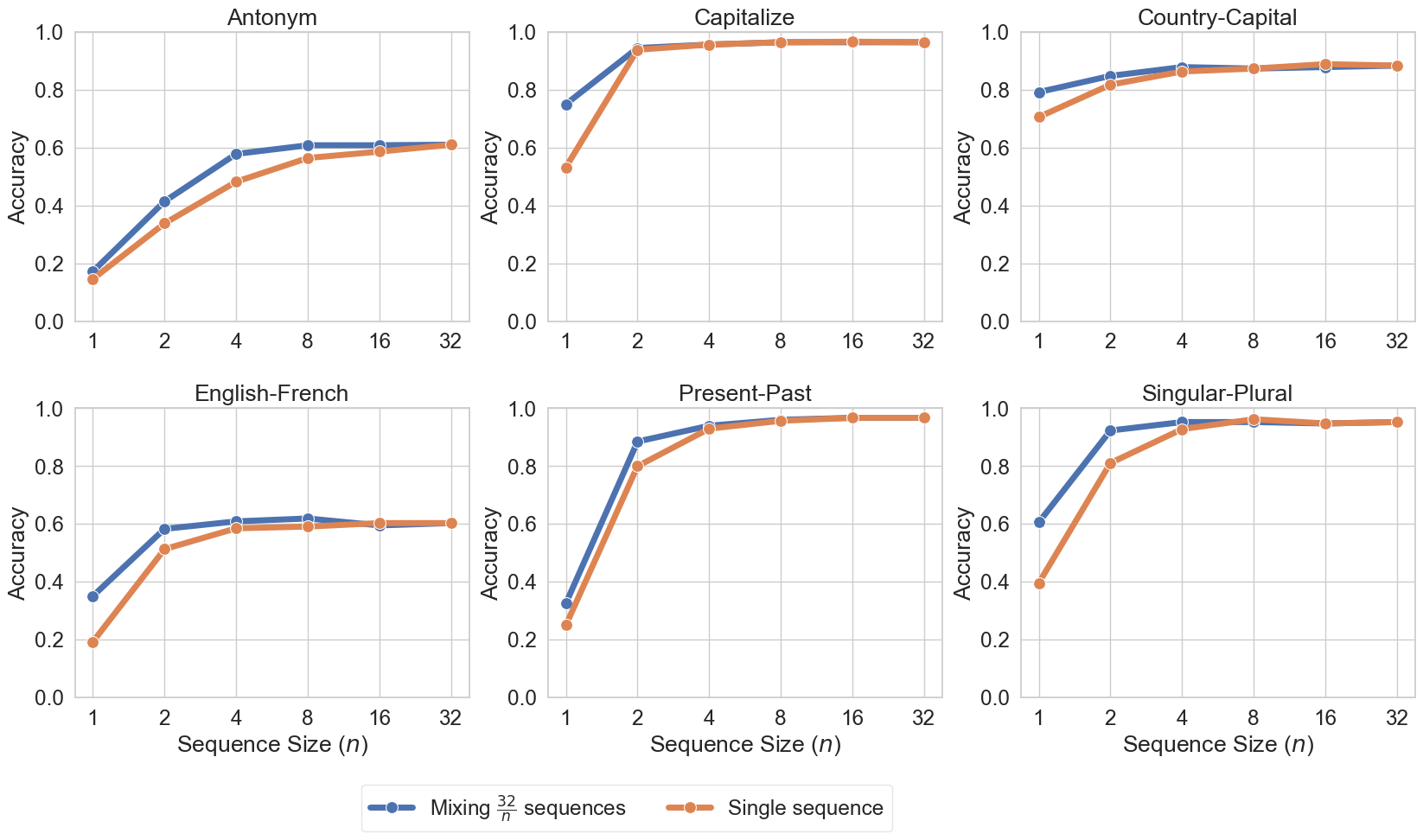}
    \caption{State mixing improves few-shot learning performance. The x-axis represents the number of examples in the processed sequence.} 
    \label{fig:mix_appendix}
\end{figure}

\begin{figure}
\centering
    \includegraphics[width=\textwidth]{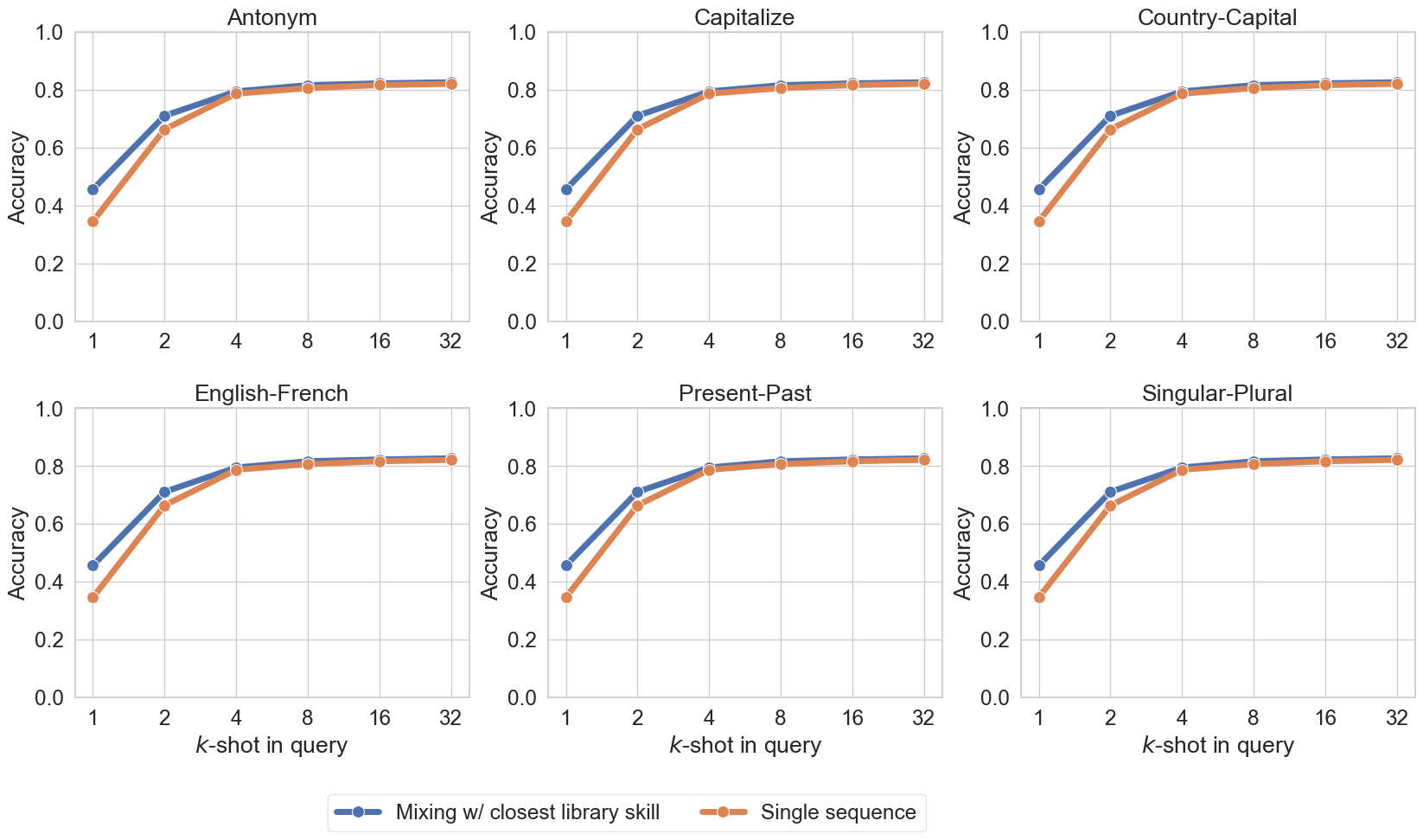}
    \caption{State retrieval and mixing improves few-shot learning performance. (Right) The x-axis represents the number of examples observed in the query state.} 
    \label{fig:retrieve_appendix}
\end{figure}

Due to limited space, the results presented in Figure \ref{fig:icl} are aggregated across tasks. Per-task results can be found in Figures \ref{fig:mix_appendix} and \ref{fig:retrieve_appendix}.

\end{document}